\documentclass[10pt,twocolumn,letterpaper]{article}

\usepackage{iccv}

\usepackage{iccv}
\usepackage{times}
\usepackage{epsfig}
\usepackage{graphicx}
\usepackage{amsmath}
\usepackage{amssymb}
\usepackage{booktabs}
\usepackage[dvipsnames]{xcolor}
\usepackage{soul}
\usepackage{float}
\usepackage{bbding}
\usepackage[nocompress]{cite}
\usepackage[inline]{enumitem}
\usepackage[
    protrusion=true,
    activate={true,nocompatibility},
    final,
    tracking=true,
    kerning=true,
    spacing=true]{microtype}
\SetTracking{encoding={*}, shape=sc}{40}
\microtypecontext{spacing=nonfrench}
\usepackage{xspace}
\usepackage{booktabs}
\usepackage{multirow}
\usepackage[accsupp]{axessibility}  
\iccvfinalcopy 

\definecolor{citecolor}{HTML}{0071bc}
\usepackage[breaklinks=true,colorlinks,citecolor=citecolor,bookmarks=false]{hyperref}

\newcommand{\ours}{$\textit{Scale-MAE}$\xspace}

\usepackage[capitalize]{cleveref}
\crefname{section}{Sec.}{Secs.}
\Crefname{section}{Section}{Sections}
\Crefname{table}{Table}{Tables}
\crefname{table}{Tab.}{Tabs.}

\def\iccvPaperID{6763} 

\ificcvfinal\fi

\begin{document}

\title{Scale-MAE: A Scale-Aware Masked Autoencoder for Multiscale Geospatial Representation Learning}
\author{Colorado J Reed\textsuperscript{1,2}\thanks{Denotes co-first authorship. Co-first authors will prioritize their names on their resumes/websites.},
Ritwik Gupta\textsuperscript{1}\footnotemark[1],
Shufan Li\textsuperscript{1}\footnotemark[1],
\\
Sarah Brockman\textsuperscript{3},
Christopher Funk\textsuperscript{3},
Brian Clipp\textsuperscript{3}, \\
Kurt Keutzer\textsuperscript{1},
Salvatore Candido\textsuperscript{2},
Matt Uyttendaele\textsuperscript{2},
Trevor Darrell\textsuperscript{1}
\\ \\ \textsuperscript{1}Berkeley AI Research; \textsuperscript{2}Meta AI, FAIR; \textsuperscript{3}Kitware Inc. \\
{\small correspondence to }{\tt\small ritwikgupta@berkeley.edu}
}
\maketitle

\begin{abstract}
Large, pretrained models are commonly finetuned with imagery that is heavily augmented to mimic different conditions and scales, with the resulting models used for various tasks with imagery from a range of spatial scales. Such models overlook scale-specific information in the data for scale-dependent domains, such as remote sensing. In this paper, we present \ours, a pretraining method that explicitly learns relationships between data at different, known scales throughout the pretraining process. \ours pretrains a network by masking an input image at a known input scale, where the area of the Earth covered by the image determines the scale of the ViT positional encoding, not the image resolution. \ours encodes the masked image with a standard ViT backbone, and then decodes the masked image through a bandpass filter to reconstruct low/high frequency images at lower/higher scales. We find that tasking the network with reconstructing both low/high frequency images leads to robust multiscale representations for remote sensing imagery. \ours achieves an average of a $2.4 - 5.6\%$ non-parametric kNN classification improvement across eight remote sensing datasets compared to current state-of-the-art and obtains a $0.9$ mIoU to $1.7$ mIoU improvement on the SpaceNet building segmentation transfer task for a range of evaluation scales. 

\end{abstract}

\section{Introduction} 
Remote sensing data is captured from satellites and planes through a mixture of sensors, processing pipelines, and viewing geometries. 
Depending on the composition and relative geometry of the sensor to the Earth, each image's Ground Sample Distance (GSD - the physical distance between two adjacent pixels in an image) can vary from 0.3m to 1km, so a 100x100 pixel image could span anywhere from an Olympic-size swimming pool (900 m\textsuperscript{2}) to almost the entire country of Jamaica (10,000 km\textsuperscript{2}).
The data within each image, and the corresponding objects and points of interest, can therefore vary across wide spatial ranges.
Data from these multiscale sensors provide critical and complementary information for various operational and research applications in areas such as atmospheric, hydrologic, agricultural, and environmental monitoring~\cite{tellmanGlobalFloodObservation2021, moranSnowpackEstimationKey2022}.

\begin{figure}[t]
    \centering
    \includegraphics[width=\columnwidth]{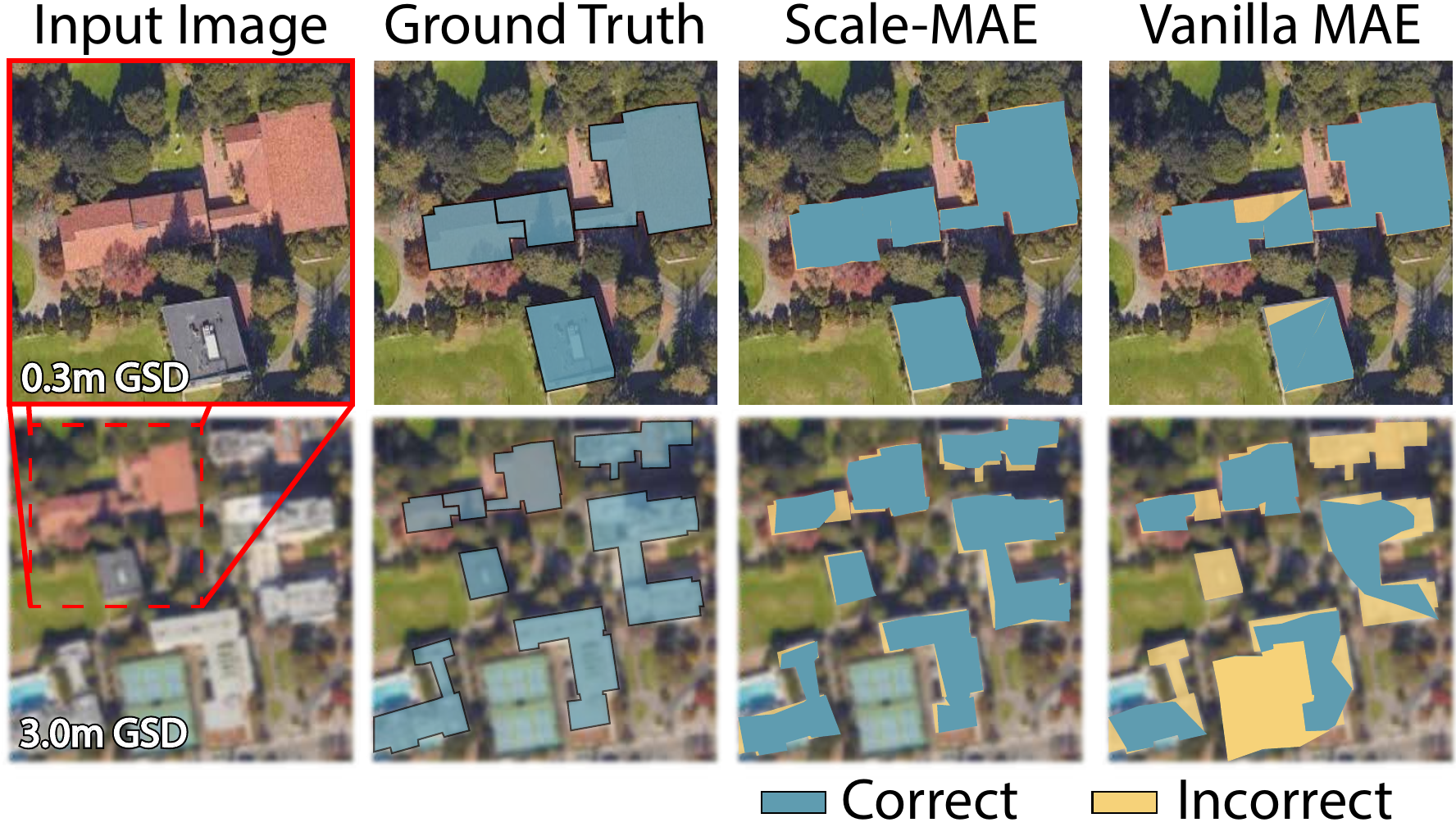}
    \caption{\textbf{\ours learns better representations for multiscale tasks compared to vanilla MAE.} (Column 1) The top image spans an area at 0.3m GSD and the bottom image shows the same region at a coarser GSD. (Columns 2-4) The following columns show a ground truth building segmentation, \ours segmentation from a finetuned UperNet, and segmentation from an analogously finetuned UperNet from a vanilla MAE, respectively. \ours demonstrates better performance across images at both scales. See the supplementary material for more examples.}
    \label{fig:teaser}
\end{figure}

Few modern computer vision methods have explicitly addressed multiscale remote sensing imagery~\cite{kowaleczko2022mus2}.
\begin{figure*}[ht]
    \centering
    \includegraphics[width=\textwidth]{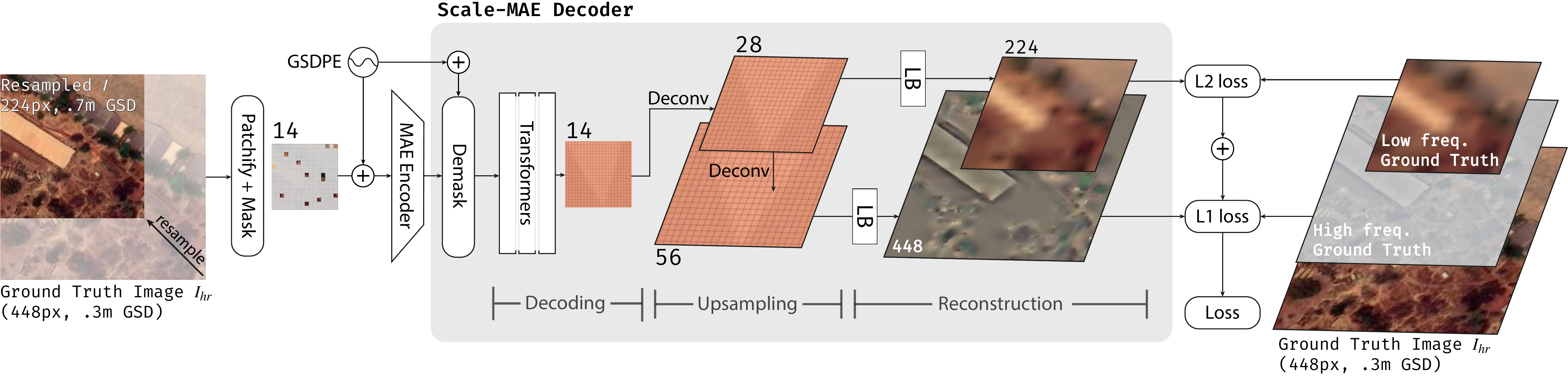}
    \caption{\textbf{\ours employs the Masked Autoencoder framework.} An input image is patchified and masked before being passed into an MAE encoder. A Ground Sample Distance Positional Encoding (GSDPE) is added to the encoder input, which scales the positional encodings to the area of ground covered. The \ours decoders has three stages: (1) Decoding, which uses a smaller number of transformer layers than MAE to decode the encoded values (2) Upsampling, which progressively deconvolves the decoded feature map to a larger size before being passed through the Laplacian Blocks (abbreviated LB, see \Cref{sec:scale-mae}), (3) Reconstruction, which then reconstructs low and high frequency features at different scales. These outputs are used to compute an aggregate loss with ground truth low and high frequency features, where following super resolution literature~\cite{anwar2020deep}, an L1 loss is used for high frequency output to better reconstruct edges and an L2 loss is used for low frequency output to better reconstruct average values.}
    \label{fig:explainer}
\end{figure*}
Nevertheless, the remote sensing vision community has increasingly used large, pretrained models~\cite{fuller2022satvit,congSatMAEPretrainingTransformers2022}, where such applications finetune a pretrained model for a single source of data at a specific scale~\cite{fuller2022satvit,congSatMAEPretrainingTransformers2022,gao2022general,ibanez2022masked,liu2022band}.
In this paper we present \ours, a masked reconstruction model that explicitly learns relationships between data at different, known scales throughout the pretraining process. 
By leveraging this information, \ours produces a pretrained model that performs better across a wide range of GSDs and tasks.

Masked Autoencoders~\cite{heMaskedAutoencodersAre2021} offer self-supervised learning without explicit augmentations. A standard Masked Autoencoder resizes/crops an image, masks the majority of the transformed image, and then tasks a Vision Transformer (ViT) based autoencoder with embedding the unmasked components.
A decoding ViT then decodes the full image from these learned embeddings, where the decoder is later discarded and the encoder is used to produce representations for an unmasked input image.

Existing MAE-based pretraining approaches fail to generalize across domains with images at multiple scales.
\ours (\Cref{fig:teaser}) overcomes this through a GSD-based positional encoding derived from the land area covered in the image. This informs the ViT of both the position \textit{and scale} of the input image.
\ours also uses a Laplacian-pyramid decoder to encourage the network to learn multiscale representations. The embeddings are decoded to two images containing low and residual high frequency information, respectively -- see \Cref{fig:explainer}.
As we discuss in \Cref{sec:scale-mae}, this structure allows the ViT decoder to use fewer parameters than MAE while still producing strong representations across multiple scales.

We show that \ours leads to better performing, more robust multiscale representations than both a standard MAE and a recently proposed, state-of-the-art MAEs SatMAE~\cite{congSatMAEPretrainingTransformers2022} and ConvMAE~\cite{gaoConvMAEMaskedConvolution2022} across remote sensing datasets with a variety of scale and resolution characteristics. To the best of our knowledge \ours is the first self-supervised MAE to include scale-aware positional encoding and Laplacian pyramids. In our experiments, \ours achieves an average of a $5.6\%$ nonparametric kNN classification improvement across eight remote sensing datasets compared to current state-of-the-art in addition to a $0.9$ mIoU to $1.7$ mIoU improvement on the SpaceNet building segmentation transfer task for a range of evaluation scales (see \Cref{fig:teaser}).

\section{Related Work} 
\paragraph{Representation learning and the Masked Autoencoder.}
Representation learning aims to extract meaningful, intrinsic features from data for downstream use~\cite{bengio2013representation}.
In practice, this often entails pretraining a deep network so that a lightweight learning routine can then finetune it for a particular downstream task, see~\cite{devlin2018bert, donahue2014decaf, erhan2010does,goodfellow2016deep, he2020momentum, henaff2020data, lecun2015deep, radford2018improving, zeiler2014visualizing}. 
The Masked Autoencoder (MAE) is a recent state-of-the-art self-supervised representation learning method in computer vision that pretrains a ViT encoder by masking an image, feeding the unmasked portion into a transformer-based encoder, and then tasking the decoder with reconstructing the input image~\cite{heMaskedAutoencodersAre2021}.  MAEs fail to leverage scale information in scale-dependent domains as they are often reliant on absolute or relative positional encodings. To the best of our knowledge, \ours is the first MAE-based self-supervised learning method to incorporate a scale-variant positional encoding.

\paragraph{Remote Sensing Representation Learning}

Neumann~\etal~\cite{Neumann2019IndomainRL} were one of the first to exhaustively share results on existing representation learning and semi-supervised learning techniques for remote sensing imagery.
Gao~\etal~\cite{gao2022general} demonstrated the effectiveness of MAE pretraining for remote sensing image classification.
Ayush~\etal~\cite{ayushGeographyAwareSelfSupervisedLearning2021} leveraged the metadata from remote sensing images via spatially aligned but temporally separated images as positive pairs for contrastive learning and predicted the latitude and longitude as pretext tasks.
Gupta~\etal~\cite{guptaAcceleratingUkraineIntelligence2022} demonstrated the use of MAEs as a pretraining approach for passive and active remote sensing imagery. Their method introduced flexible ``adapters" which could be used interchangeably with an encoder for a set of input imagery modes.
Cong~\etal~\cite{congSatMAEPretrainingTransformers2022} introduced the SatMAE, which used temporal and spectral metadata in a positional encoding to encode spatio-temporal relationships in data. The temporal data contains the year, month, and hour enabling understanding of long-term change with the year, weather information from the month, and hour information for the time of day.
Further Liu~\etal~\cite{liu2022band} and Iba{\~n}ez~\etal~\cite{ibanez2022masked} have shown that MAE architectures can be used for band selection in hyperspectral remote sensing images, significantly reducing data redundancy while maintaining high classification accuracy. \ours leverages inherent absolute scale information information present in scale-dependent domains as a way to learn robust, multiscale features that reduce data usage downstream.

\paragraph{Super-resolution}

Super-resolution has proven effective in improving accuracy within remote sensing images due to the extremely small size of objects within the image~\cite{shermeyer2019effects}. 
Previous works have aimed to learn continuous implicit representations for images at arbitrary resolutions to aid the super-resolution task. These representations are used to upsample the images either to specific scales~\cite{ledig2017photo} or to arbitrary resolutions~\cite{xu2021ultrasr, chen2021learning, hu2019magnification}. Most super-resolution work aims to increase the resolution of the input image, whereas \ours produces both higher \textit{and lower} resolution images. There is some work on super-resolution for satellite imagery, but much of this work is focused on synthetically creating high-resolution datasets for use with models trained specifically for high-resolution data \cite{he2021spatial, kowaleczko2022mus2}. \ours, however, utilizes super-resolution as a means to obtain multiscale representations during pretraining.

\paragraph{Multiscale Features} Because images can contain objects of many different pixel resolutions, the vision community has proposed many methods to extract multiscale features. 
These include spatial pyramids \cite{lazebnikBagsFeaturesSpatial2006, burtLaplacianPyramidCompact1983, rosenfeldEdgeCurveDetection1971, koenderinkStructureImages1984} and dense sampling of windows \cite{yanSpatialPyramidsNew2012a, jiRegionBasedSpatialSampling2013, yanImageClassificationDensely2015}. 
These approaches have been combined by methods such as \cite{felzenszwalbDiscriminativelyTrainedMultiscale2008}, in which dense histogram-of-gradient features are computed for each feature pyramid level. 
Rather than using classical computer vision techniques to extract multiscale features, convolutional neural networks have been used to build deep multiscale features. 
CNNs with subsampling layers inherently build feature pyramids, a property exploited explicitly by models such as the Feature Pyramid Network and the Single-Shot Detector, amongst others~\cite{liuSSDSingleShot2016a, linFeaturePyramidNetworks2017, ghiasiLaplacianPyramidReconstruction2016}.
Recently, this multiscale idea has been extended to vision transformers by~\cite{fan2021multiscale}, who show that this architecture improves various video recognition and image classification tasks, as well as in \cite{gaoConvMAEMaskedConvolution2022,zhangPointM2AEMultiscaleMasked2022} which proposes various hybrid CNN-MAE architectures that yield multiscale features during MAE pretraining.
Different from these works, \ours uses a Laplacian pyramid decoder as a way to force an encoder to learn multiscale features with the ViT architecture.

\section{Scale-MAE}
\label{sec:scale-mae}
This section describes the \ours pretraining framework as illustrated in \Cref{fig:explainer}.
\ours is a self-supervised pretraining framework based on the Masked Autoencoder (MAE) \cite{heMaskedAutoencodersAre2021}.
\ours makes two contributions to the MAE framework. Standard MAE-based methods use absolute or relative positional encodings to inform the ViT of the position of the unmasked components, where an image at resolution $r$ will have the same positional encodings regardless of the image content.
\ours introduces the Ground Sample Distance (GSD) based positional encoding that scales in proportion to the area of land in an image, regardless of the resolution of the image. 
In addition, \ours introduces the Laplacian-pyramid decoder to the MAE framework to encourage the network to learn multiscale representations. Embeddings from a ViT encoder are decoded to a lower resolution image that captures the lower frequency information and a higher resolution image that captures the high-frequency information.
We formalize \ours in the following subsections by first specifying the necessary MAE background, describing the GSD-based positional encoding, and then explaining the Laplacian-pyramid decoder.

\begin{figure}
    \centering
    \includegraphics[width=\columnwidth]{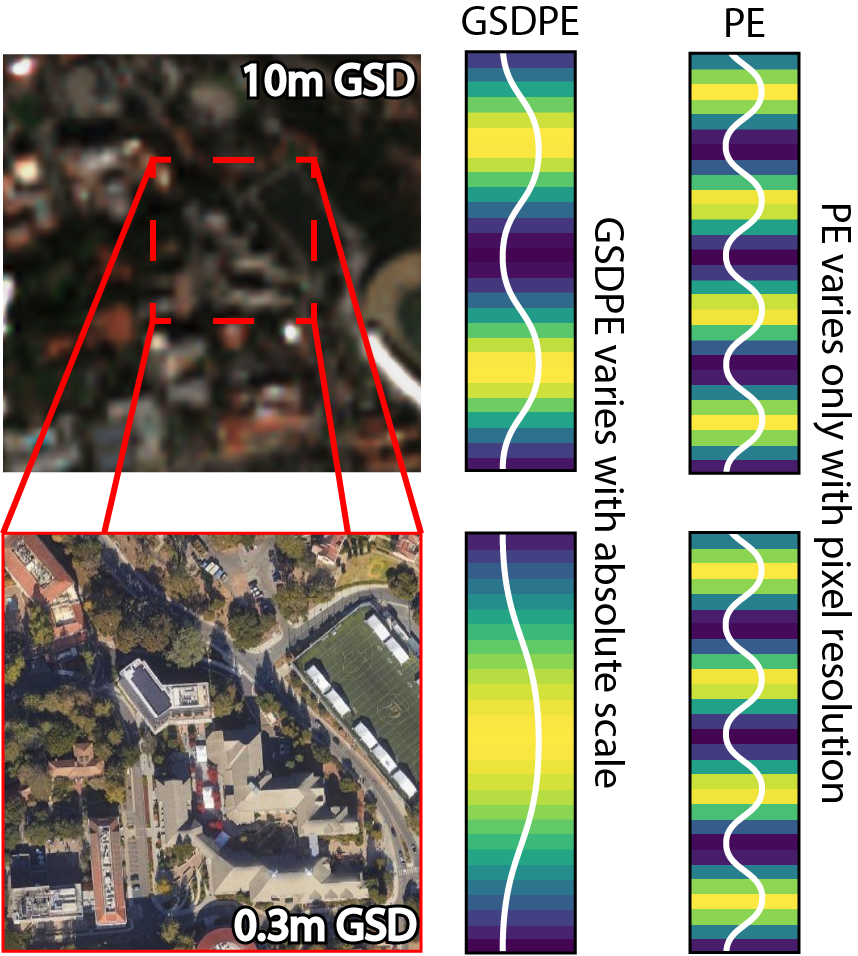}
    \caption{\textbf{Ground Sample Distance Positional Encoding (GSDPE).} (Left) Input images at the same pixel resolution but different GSDs are shown. The image on the bottom is a subset of the image on the top. (Center) This overlap in location, albeit at a different resolution, is reflected in the GSDPE. The finer image with smaller spatial extent is represented by a corresponding subsection of the overall sine wave on the bottom. (Right) A standard positional encoding is strictly dependent on the image resolution and uses the same embedding for both. The colors behind the sine waves show the intensity and quantization of the encoding.}
    \label{fig:gsdpe}
\end{figure}

\paragraph{Setup}
Let $I \in \mathbb{R}^{H \times W \times C}$ represent an input image of height $H$, width $W$, and $C$ channels.
The MAE \textit{patchifies} $I$ into a sequence $S$ of independent patches of height and width $P$ pixels, where each of the $N_p$ patches, $s\in S$ has dimension $s\in \mathbb{R}^{P^2C}$.
A fraction, $m$, of the patches are then removed and the remaining patches are then passed through a projection function 
(e.g.,~a linear layer) to project the patches $S$ into $D$ dimensions, 
$f_E: \mathbb{R}^{P^2C} \rightarrow \mathbb{R}^D$,
to obtain embedded patches $S_E = f_E(S)$. 
An $\mathbb{R}^2$ positional encoding vector, is then added to the embedded patches with
\begin{align}
\small
v_{x}(pos, 2i)&=\sin \frac{pos}{10000^{\frac{2i}{D}}} \\
v_{y}(pos, 2i+1)&=\cos \frac{pos}{10000^{\frac{2i}{D}}}
\label{eq:pos_enc}
\end{align} 
where $pos$ is the position of the patch along the given axis and $i$ is the feature index (visualized in \Cref{fig:gsdpe}), exactly as introduced in \cite{vaswaniAttentionAllYou2017}.
These positional encodings are then concatenated and added to the embedded patches, which are then fed into a ViT encoder. 
After the encoder, the removed $m$ patches are then placed back into their original location in the sequence of patches where a \textit{learned mask token} represents the masked patches that were not encoded. 
Another positional encoding vector is added to all patches and a sequence of transformer blocks decodes these patches to form the original input image, which is used as the learning target.

\paragraph{Input} \ours performs a super resolution reconstruction, where the input image $I$ is downsampled from a higher resolution image $I_{\text{hr}}$ at the ground truth GSD. Instead of targeting the input image, \ours targets high frequency and low frequency components of $I_{\text{hr}}$, which is common in Laplacian pyramid super resolution models~\cite{yang2019deep}, where the high frequency component is at the same resolution as the ground truth image $I_{\text{hr}}$ and the low frequency component is at the same resolution as the input image $I$, as shown in \Cref{fig:explainer}.
Following many works in super resolution\cite{yang2019deep}, the low frequency target image is obtained by interpolating $I_{\text{hr}}$ to a much lower resolution, $r_{\text{low}}$ and then interpolating to the same resolution as the input image $I$.
The high frequency target image is obtained by downsampling $I_{\text{hr}}$ to another lower resolution $r_{\text{high-low}}$, and then upsampling to the same resolution as the ground truth image $I_{hr}$ and subtracting this image $I_{\text{hf}} = I_{\text{hr}} - I_{\text{high-low}}$. 
The supplementary material provide more information on the upsampling/downsampling methodology.
The key components for \ours are described next.

\paragraph{GSD Positional Encoding}
Images from scale-dependent domains have a metric which defines the absolute scale for the image. This metric has different names across domains and is referred to as the Ground Sample Distance (GSD) in remote sensing.
The GSD is critical to understanding, conceptually, the kinds of features that will be available in an image. 
An image with finer GSD (lower number) will have higher frequency details than an image with coarser GSD (high number). 
Models are generally unaware of absolute scale when learning over a set of data. Specifically, even if they implicitly learn that all images in a dataset share a varying resolution from input-space augmentations, then these models do not explicitly condition on the GSDs encountered in unseen data.

We extend the positional encoding from \Cref{eq:pos_enc} to include GSD by scaling the positional encoding relative to the land area covered in an image as depicted in \Cref{fig:gsdpe} and mathematically: 
\begin{align}
\small
    v_{gsd,x}(pos, 2i) &=\sin \textcolor{blue}{\frac{g}{G}}\frac{pos}{10000^{\frac{2i}{D}}} \\
    v_{gsd,y}(pos, 2i+1) &=\cos \textcolor{blue}{\frac{g}{G}}\frac{pos}{10000^{\frac{2i}{D}}}
\label{eq:gsd_pos_enc}
\end{align} 
where $g$ is the GSD of the image and $G$ is a reference GSD, nominally set to 1m.
Intuitively, an object imaged at a finer resolution has more pixels representing it. When imaging the same object at a coarser resolution, those pixels must map to fewer pixels. In \Cref{eq:gsd_pos_enc}, we interpolate the positional encoding by a factor of $\frac{G}{g}$ to account for the ordering of the coarser set of pixels. This simple idea underpins the GSD Positional Encoding, visualized in \Cref{fig:gsdpe}.

\paragraph{Scale-MAE decoder}
The standard MAE learns representations by tasking a network with reconstructing an image after masking out most of its pixels. 
While the standard MAE decoder reconstructs the input image at the same scale as its input, the objective of \ours is to learn multiscale representations. 
We draw on works from progressive super-resolution such as \cite{wangFullyProgressiveApproach2018}, that learn a high resolution, high frequency image and a lower resolution low frequency image, that when combined together, yield the input image at a higher resolution.

The \ours introduces a novel decoder which decodes to multiple scales with a progressive Laplacian decoder architecture, replacing the traditional MAE ``decoder", which is really a Transfomer encoder.
This architecture consists of three stages: decoding, upsampling, and reconstruction, which are shown in \Cref{fig:explainer} and detailed below.

\textit{Decoding} follows the standard MAE decoder where following the encoder, the removed $m$ patches are then placed back into their original location in the sequence of patches where a \textit{learned mask token} represents the masked patches that were not encoded, a positional encoding is added, and then a series of transformer layers decode all patches.
In contrast to the standard MAE decoder, the \ours decoder uses fewer transformer layers (e.g.~3 layers instead of 8), which reduces the parameter complexity as quantified in \Cref{sec:discussion}.
The output of these layers is then fed into the upsampling stage.

\textit{Upsampling} 
The latent feature maps from the decoding stage are progressively upsampled to 2x and 4x resolution using deconvolution blocks, where the first deconvolution is 2x2 with stride 2 that outputs a feature map at 2x the input resolution (28 in \Cref{fig:explainer}), followed by a LayerNorm and GELU, and then another 2x2 deconvolution layer that outputs a feature maps at 2x the previous resolution (56 in \Cref{fig:explainer}). See the supplementary material for a full architectural diagram.

\textit{Reconstruction}
After having been upsampled, the lower resolution and higher resolution feature maps are passed into \textit{Laplacian Blocks} (LBs in \Cref{fig:explainer}) that reconstruct high and low resolution images for the high and low frequency reconstruction, respectively. Architecturally, the Laplacian Blocks consist of a sequence of three sub-blocks: a Laplacian Feature Mapping Block, a Laplacian Upsample Block, and a Laplacian Pyramid Reconstruction Block.
The Feature Mapping Block is used to project features within a particular layer of the Laplacian Pyramid back to the RGB space.  
The Laplacian Upsample Block represents a learnable upsample function that maps latent features from one layer of the Laplacian Pyramid to a higher level. 
Finally, the Laplacian Pyramid Reconstruction Block is used to reconstruct information at the different frequencies in RGB space. Following super resolution literature~\cite{anwar2020deep}, an L1 loss is used for high frequency output to better reconstruct edges and an L2 loss is used for low frequency output to better reconstruct average values.
The supplementary material has architectural diagrams for each block.

\begin{figure}
    \centering
    \includegraphics[width=\columnwidth]{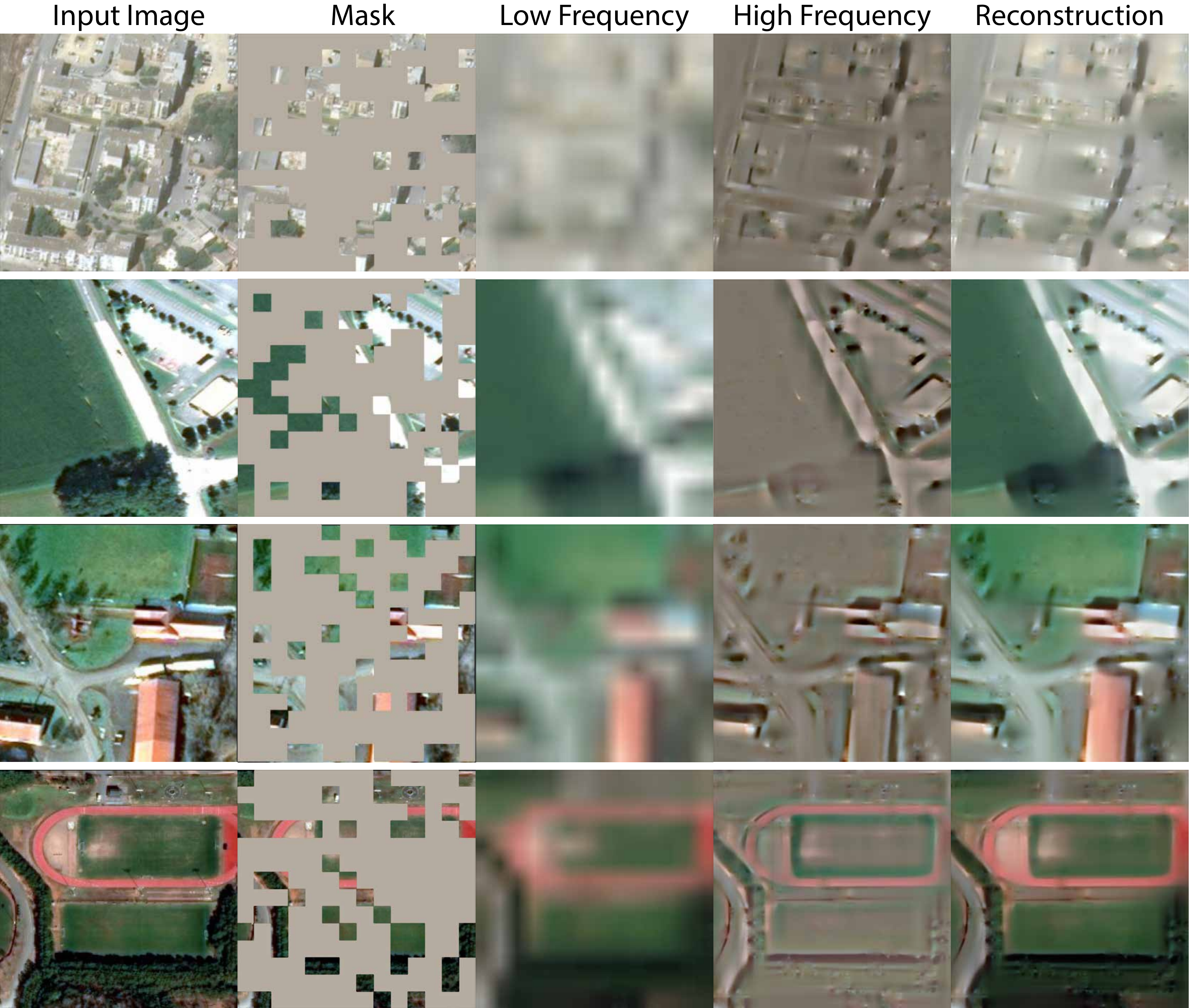}
    \caption{\textbf{\ours reconstruction.} Examples from Functional Map of the World are shown. From left to right, an input image at 224x224 resolution is shown. Its corresponding mask is visualized as well. Columns 3 and 4 show the low and high frequency produced by the \ours decoder. The last column is the reconstruction obtained from summing the low and high frequency features together.}
    \label{fig:reconstruction}
\end{figure}

\section{Experiments}
\label{sec:experiments}
\begin{figure*}
    \centering
    \includegraphics[width=\textwidth]{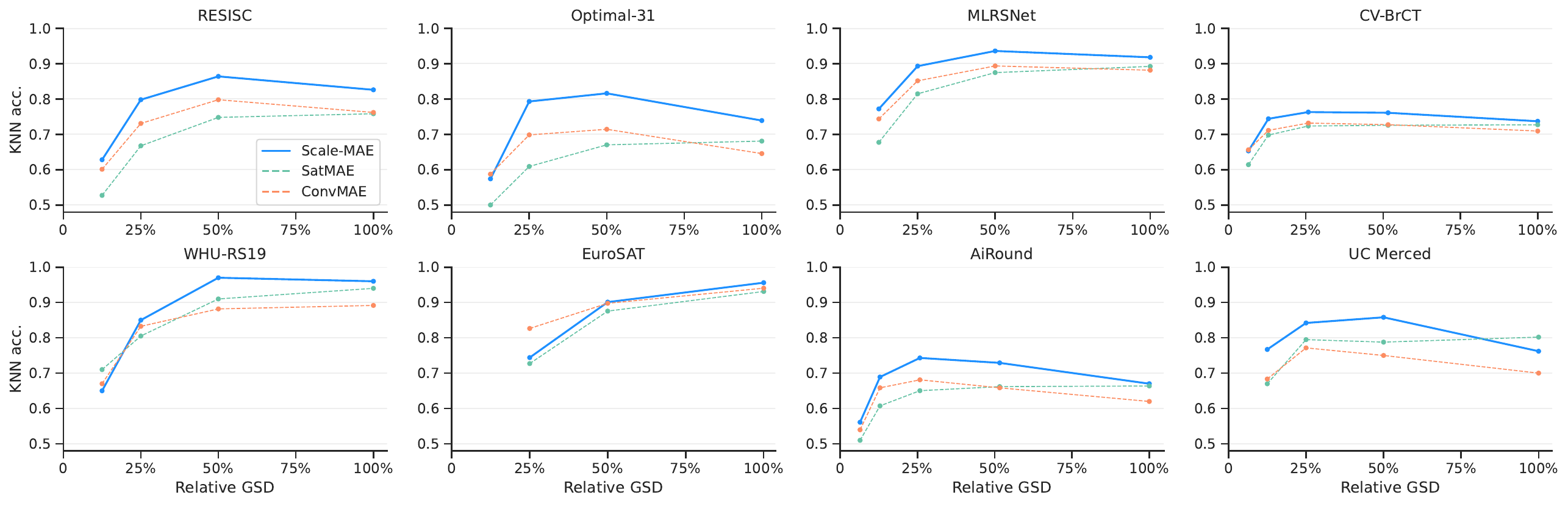}
    \caption{\textbf{Learning better representations at all scales.} \ours (blue) features perform better than state-of-the-art. We evaluate kNN accuracy on eight datasets with a large variance in GSD. \ours consistently produces better results at coarser resolutions. In addition to using evaluation datasets at different GSDs, to further test the multiscale representations, we create multiple test sets for each dataset in which we downsampled the full resolution validation set to coarser GSDs at fixed percentages:  $X_{val}^{G\%}, G \in \{12.5, 25, 50, 100\}$, where EuroSat does not include the 12.5\% because the images are at a resolution of 64px, our patch size is 16px, and an input image of 8px is too small.}
    \label{fig:knn-eval}
\end{figure*}

We investigate the quality of representations learned from \ours pretraining through a set of experiments that explore their robustness to scale as well as their transfer performance to additional tasks. 
First, we present our main experiments in \Cref{sec:main experiments} and compare with SatMAE~\cite{congSatMAEPretrainingTransformers2022}, a current state-of-the-art MAE for remote sensing imagery, ConvMAE~\cite{gaoConvMAEMaskedConvolution2022}, a state-of-the-art multiscale MAE, as well as several other approaches detailed throughout. The exact implementation of \ours for the main experiments was determined through a set of ablation experiments presented in \Cref{sec:ablations}.

We pretrain a ViT-Large model with \ours using the Functional Map of the World (FMoW)~\cite{christieFunctionalMapWorld2018} RGB training set, which consists of $363.6$k images of varying image resolution and GSD, for 800 epochs. 
The initial higher resolution image $I_{\text{hr}}$ is taken as a random 448px$^2$ crop of the input image, and the input image $I$ is then a downsampled 224px$^2$ from $I_{\text{hr}}$.
The low frequency groundtruth is obtained by downscaling $I_{\text{hr}}$ to 14px$^2$ and then upscaling to 224px$^2$, while the high frequency groundtruth is obtained by downscaling $I_{\text{hr}}$ to 56px$^2$ and then upscaling to 448px$^2$ and subtracting this image from $I_{\text{hr}}$.

\Cref{fig:reconstruction} shows examples of the masked input, low resolution/frequency, high resolution/frequency, and combined reconstruction of FMoW images during training. The low resolution/frequency images capture color gradients and landscapes, while the residual high resolution/frequency images capture object edges, roads, and building outlines.


\subsection{Representation Quality}
\label{sec:main experiments}
We evaluate the quality of representations from \ours by freezing the encoder and performing a nonparametric k-nearest-neighbor (kNN) classification with eight different remote sensing imagery classification datasets with different GSDs, none of which were encountered during pretraining.
The kNN classifier operates by encoding all train and validation instances, where each embedded instance in the validation set computes the cosine distance with every other embedded instance in the training set. The instance is classified correctly if the majority of its k-nearest-neighbors are in the same class as the validation instance, and incorrectly if they are in any other.

The reasoning behind the kNN classifier evaluation is that a strong pretrained network will output semantically grouped representation for unseen data of the same class. This evaluation for the quality of representations occurs in other notable works~\cite{caronEmergingPropertiesSelfSupervised2021, chenExploringSimpleSiamese2021, wuUnsupervisedFeatureLearning2018}.
In addition to using evaluation datasets at different GSDs, to further test the multiscale representations, we create multiple test sets for each dataset. Since we cannot synthesize data at a finer GSD than the provided ground truth, we only downsample the full resolution validation set to coarser GSDs at fixed percentages:  $X_{val}^{G\%}, G \in \{12.5, 25, 50, 100\}$.

\begin{table}[t]
\centering
\begin{tabular}{c*{4}{ccc}}
~ & \multicolumn{3}{c}{Average Accuracy (\%)} \\\cline{2-4}
\multicolumn{1}{c}{\textbf{Dataset}} & \multicolumn{1}{c}{\textbf{\ours}} & \multicolumn{1}{c}{\textbf{SatMAE}} & \multicolumn{1}{c}{\textbf{ConvMAE}}\\ \toprule
AiRound                     & \textbf{63.2} & 57.8 & 59.7                     \\\hline
CV-BrCT                     & \textbf{69.7} & 66.2 & 68.4                     \\\hline
EuroSAT                     & 86.7 & 84.4 & \textbf{88.8}                     \\\hline
MLRSNet                     & \textbf{81.7} & 75.0 & 79.5                     \\\hline
OPTIMAL-31                  & \textbf{65.5} & 55.7 & 61.7                     \\\hline
RESISC                      & \textbf{70.0} & 61.0 & 67.0                     \\\hline
UC Merced                   & \textbf{75.0} & 69.8 & 70.0                     \\\hline
WHU-RS19                    & \textbf{79.5} & 78.5 & 77.0                     \\\bottomrule
\end{tabular}
\vspace{0.3cm}
\caption{\ours performs better, across all GSDs (as in \Cref{fig:knn-eval}), for all datasets we experimented with compared to SatMAE. The average improvement across all datasets for \ours compared to SatMAE is 5.6\% and 2.4\% compared to ConvMAE with ViT-Large backbones.}
\label{tab:knn-per-dataset}
\end{table}

Our analysis uses eight different land-use classification datasets: RESISC-45 \cite{chengRemoteSensingImage2017}, the UC Merced Land Use Dataset \cite{yangBagofvisualwordsSpatialExtensions2010}, AiRound and CV-BrCT \cite{machadoAiRoundCVBrCTNovel2021}, MLRSNet \cite{qiMLRSNetMultilabelHigh2020}, EuroSAT \cite{helberIntroducingEurosatNovel2018}, Optimal-31 \cite{wangSceneClassificationRecurrent2019}, WHU-RS19 \cite{daiSatelliteImageClassification2011}, SpaceNet v1 and v2 \cite{vanettenSpaceNetRemoteSensing2019}, and Functional Map of the World \cite{christieFunctionalMapWorld2018}. 
The datasets used span a wide range of GSDs, e.g.,~MLRSNet consists of data captured from aerial platforms with 0.1m GSD, while RESISC45 has imagery from medium-resolution satellites at \textgreater30m GSD. 
In some cases, the datasets present imagery at mixed GSDs which are not specified, in which case we assume an approximate constant GSD: see the supplementary material for all details. Furthermore, we provide an expanded set of experiments with linear probing and finetuning in the supplementary material.

We run kNN classification with $k=20$. \Cref{fig:knn-eval} shows that \ours outperforms SatMAE and ConvMAE across GSD scales in the different evaluation datasets and across relative GSD scales within individual datasets. For example, the UC Merced has a GSD of 0.3m, but evaluating at scales [12.5\%,100\%] provides an artificial GSD range of [0.3m, 2.4m]. On this example, we see that \ours provides the largest performance gap at the 2.4m GSD, with similar performance at 0.3m.

Across all other evaluation datasets and wider range of GSDs, \ours outperforms SatMAE and ConvMAE, where \ours outperforms both methods by a larger gap as the GSD increasingly varies from the original GSD, indicating that \ours learns representations that are more robust to changes in scale for remote sensing imagery. We outperform SatMAE by an average of 5.6\% and ConvMAE by an average of 2.4\% across all resolutions and datasets (see \Cref{tab:knn-per-dataset}). UC Merced at 100\% of the true GSD is the only evaluation where SatMAE outperforms \ours. The supplementary material contains an extensive table demonstrating kNN classification results with varying $k$.

\paragraph{Linear probing and finetuning} We perform linear classification on the RESISC-45 and FMoW-RGB datasets. We fine-tune for 50 epochs using the same hyperparameter settings as SatMAE~\cite{congSatMAEPretrainingTransformers2022}: a base learning rate of $5 \times 10^{-3}$, a weight decay of $5 \times 10^{-3}$. We do not use temporal data for classification. For RESISC-45, we fine-tune for 100 epochs with a base learning rate of $4 \times 10^{-3}$, a weight decay of $5 \times 10^{-3}$, and a global batch size of 256 across 2 GPUs. The learning rate on the backbone is multiplied by a factor of $0.1$. We use RandomResizedCrop for augmentation. We train on 224x224 images and evaluate 256x256 images because we found evaluating at a higher scale improves the performance of all models. We report both the performance of end-to-end fine-tuning and linear probing with a frozen backbone. The linear probing setup was the same as finetuning except the learning rate was $0.1$. The results are shown in \Cref{fig:linear_prob} and \Cref{fig:linear_prob2}.
\begin{table}[htb]
  \centering
    \begin{tabular}{c c c}
    \textbf{Model} & Backbone &\textbf{Frozen/Finetune}  \\
    \hline
    \ours & Vit-Large & \textbf{89.6}/\textbf{95.7 } \\
     SatMAE\cite{congSatMAEPretrainingTransformers2022} & Vit-Large  & 88.3/94.8  \\
     ConvMAE \cite{gaoConvMAEMaskedConvolution2022} & ConvVit-Large & 81.2/95.0 \\
     MAE\cite{heMaskedAutoencodersAre2021} & Vit-Large  & 88.9/93.3 \\
    \end{tabular}
    \vspace{0.3cm}
    \caption{\textbf{Transfer classification results on RESISC-45. } Frozen indicates a linear probe and finetune is a full end-to-end finetuning of the entire model.}
    \label{fig:linear_prob}
\end{table}

\begin{table}[ht]
  \centering
    \begin{tabular}{c c c}
    \textbf{Model}  & \textbf{ Backbone} &\textbf{Top-1/Top-5}  \\
    \hline

             \ours  & ViT-Large & \textbf{77.9}/\textbf{94.3} \\
     SatMAE $\dagger$ \cite{congSatMAEPretrainingTransformers2022} &ViT-Large & 72.4/91.9  \\
     MAE \cite{heMaskedAutoencodersAre2021} &ViT-Large & 68.4/90.3  \\
              ConvMAE \cite{gaoConvMAEMaskedConvolution2022} & ConvVit-Large & 74.1/91.4 \\
     SatMAE $\ast$ \cite{congSatMAEPretrainingTransformers2022} &ViT-Large & 77.8/-  \\
     
         GASSL \cite{ayush2021geography} &ResNet-50 & 71.55/-  \\
         MoCo-V2 \cite{he2020momentum} &ResNet-50 & 64.34/- \\

    \end{tabular}
    \vspace{0.3cm}
    \caption{\textbf{Full finetuning results on FMoW-RGB. } $\dagger$: We reproduce SatMAE and ConvMAE by taking their publicly available codebases and pretraining on FMoW dataset for 800 epochs. The results differ from their reported results, but are evaluated consistently with ours. * Reports the results from the SatMAE paper~\cite{congSatMAEPretrainingTransformers2022}.}
    \label{fig:linear_prob2}
\end{table}

\paragraph{Semantic segmentation transfer} We use the SpaceNet v1 building segmentation dataset~\cite{vanettenSpaceNetRemoteSensing2019} to evaluate semantic segmentation results on contrastive and MAE-based pretraining methods. Prior methods relied on the PSANet~\cite{zhaoPSANetPointwiseSpatial2018} segmentation architecture, while \ours uses the UperNet~\cite{xiaoUnifiedPerceptualParsing2018} segmentation architecture which is more common for ViT backbones. For even comparison, we test the current state-of-the-art SatMAE and ConvMAE methods with UperNet as well. 
Results are detailed in \Cref{tab:spacenet}.

\begin{table}[h]
\centering
\begin{tabular}{c*{4}{ccc}}
\textbf{Method} & \textbf{Backbone} & \textbf{Model} & \textbf{mIoU} \\\toprule
Sup. (Scratch) & ResNet50 &PSANet & 75.6 \\\hline
GASSL \cite{ayushGeographyAwareSelfSupervisedLearning2021} & ResNet50 &PSANet & 78.5 \\\midrule[0.2em]
Sup. (Scratch)  & ViT-Large &PSANet & 74.7 \\\hline
SatMAE\cite{congSatMAEPretrainingTransformers2022} & ViT-Large &PSANet & 78.1  \\\midrule[0.2em]
Sup. (Scratch)  & ViT-Large &UperNet & 71.6 \\\hline
Vanilla MAE  & ViT-Large &UperNet & 77.9 \\\hline
SatMAE & ViT-Large &UperNet & 78.0 \\\hline
ConvMAE & ViT-Large &UperNet & 77.6 \\\midrule[0.2em]
\ours & ViT-Large &UperNet & \textbf{78.9} \\\bottomrule
\end{tabular}
\vspace{0.3cm}
\caption{Semantic segmentation results on SpaceNet v1. \ours outperforms other methods across backbone and segmentation architectures, where Sup.~(Scratch) indicates a supervised model trained from scratch (a randomly initialized network).}
\label{tab:spacenet}
\end{table}

With the same pretraining settings, \ours outperforms SatMAE by 0.9 mIoU, ConvMAE by 1.3 mIoU, and a vanilla MAE by 1.0 mIoU. \ours outperforms all other prior work, including GASSL~\cite{ayushGeographyAwareSelfSupervisedLearning2021}, which SatMAE did not outperform on the mean Intersection over Union (mIoU) metric for semantic segmentation. Particularly, \ours increases the gap in performance as the resolution of input imagery becomes coarser, highlighting the absolute scale-invariance introduced by our method.

\begin{figure}[t]
    \centering
    \includegraphics[width=\columnwidth]{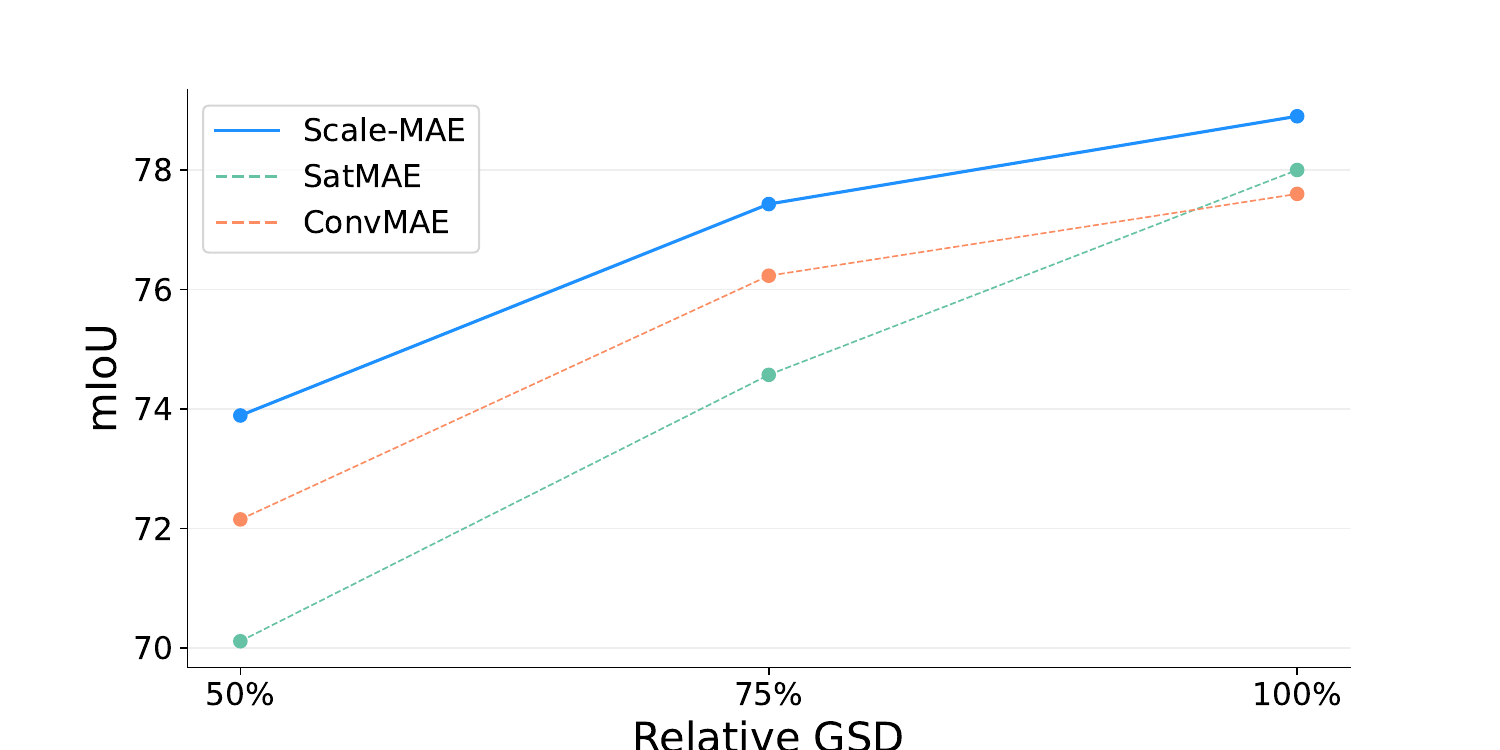}
    \caption{SpaceNet v1 evaluation across downscaled images for both \ours and SatMAE. \ours maintains higher semantic segmentation performance over SatMAE, even with images of coarser GSD.}
    \label{fig:spacenet-multires}
\end{figure}

In \Cref{fig:spacenet-multires}, we compare SpaceNet v1 evaluations across downscaled images (at 50\%, 75\%, and 100\% of the original image size) for \ours, SatMAE, and ConvMAE. Similar to the classification results, \ours maintains higher semantic segmentation performance over both methods, even with images at a coarser GSD. In fact, the performance gap grows at coarser GSDs. Compared to the next-best-performing method at the input GSD, \ours is 0.9 mIoU higher, at 75\% GSD \ours is 1.2 mIoU higher, and at 50\% \ours is 1.7 mIoU higher.

In \Cref{tab:semseg}, we further evaluate \ours, SatMAE, and ConvMAE across SpaceNet v1, SpaceNet v2~\cite{vanettenSpaceNetRemoteSensing2019}, INRIA Aerial Image~\cite{maggiori2017dataset}, and GID-15~\cite{GID2020} remote sensing datasets at native resolution. \ours outperforms both comparable methods across all benchmarks.

\begin{table}[t]
\centering
\begin{tabular}{c|c|cccc|c|c}
      & \multicolumn{1}{c|}{SN1}  & \multicolumn{4}{c|}{SN2}                                       & INR. & G15 \\
   & RI  & SH      &  VE         & PA         & KH      & -    & -    \\
   \hline
Conv. & 77.6 & 78.7          & 82.2         & 78.3         & 74.8          & 82.2      &   37.4  \\
Sat.  & 78.0 & 81.9          & 86.6          & 80.3          & 76.1          & 83.0    &  44.3   \\
Scale & \textbf{78.9 }& \textbf{82.2} & \textbf{87.4} & \textbf{81.1} & \textbf{77.1} & \textbf{84.2 }     & \textbf{46.2}
\end{tabular}
\vspace{0.3cm}
\caption{mIoU on semantic segmentation tasks. SN1/2 (SpaceNet v1/2), RI: Rio, SH: Shanghai, VE: Vegas, PA: Paris, KH: Khartoum; INR: INRIA; G15: GID-15. Conv., Sat., and Scale. are ConvMAE, SatMAE, and Scale-MAE.}
\label{tab:semseg}
\end{table}

\subsection{Ablations}
\label{sec:ablations}

\begin{table}[t]
\centering
\begin{tabular}{c*{4}{ccc}}
\textbf{Method} & \textbf{GSDPE} & \textbf{KNN 50\%} & \textbf{KNN 100\%}\\\toprule
Vanilla MAE & & 72.8 & 77.8 \\\hline
Vanilla MAE &\Checkmark  &75.4  & 78.5 \\\midrule[0.2em]
MAE + LP &  & 75.3 & 79.6 \\\hline
\ours &\Checkmark & \textbf{78.1} & \textbf{80.7} \\\bottomrule
\end{tabular}
\vspace{0.3cm}
\caption{Ablation results indicating the importance of GSDPE as determined by a KNN classification on RESISC-45 at a relative GSD of 50\% and 100\% of its native GSD. Using the GSDPE leads to better performance for both \ours and the Vanilla MAE. MAE + LP denotes the vanilla MAE with the addition of our progressive Laplacian decoder.}
\label{tab:gsdpe}
\end{table}

We ablate the key components of the \ours pretraining framework.
For these experiments, we use a lightweight pretraining setting, where we pretrain for 300 epochs on FMoW (rather than 800) and use a ViT-Base encoder (rather than ViT-Large), and evaluate using a kNN evaluation on RESISC-45 at 100\% and 50\% of its native GSD.
The key contributions that we ablate are as follows: the GSD positional encoder in \Cref{tab:gsdpe}, in which we find that the GSD postional encoder benefits both \ours and Vanilla MAE across resolutions. In \Cref{tab:numlay}, we see that the number of transformer layers can be reduced from 8 to 3 compared to a Vanilla MAE, which results in a performance improvement. The standard masking rate of $75\%$ still appears optimal for \ours according to the results in \Cref{tab:mask}.

\begin{table}[h]
\centering
\begin{tabular}{c*{4}{ccc}}
 \textbf{Mask Rate} & \textbf{KNN 50\%} & \textbf{KNN 100\%}\\\toprule
70\%& 77.3  & 79.3 \\\hline
75\%& \textbf{78.1} & \textbf{80.7} \\\hline
80\%& \textbf{78.1}  & 79.9 \\\hline
\end{tabular}
\vspace{0.3cm}
\caption{Ablation results indicating that a 75\% mask rate is optimal as determined by a KNN classification on RESISC-45 at a relative GSD of 50\% and 100\% of its native GSD.}
\label{tab:mask}
\end{table}

In \Cref{tab:resolution} we ablate the necessity of the low and high resolution reconstructions. Specifically, 
we test reconstructing the low resolution image only, the high resolution image, and a combined image (rather than independent low/high reconstructions). In this case, when the high resolution component is reconstructed, we do not use the low-resolution residual, but rather, directly reconstruct the high resolution result. The ``Combined'' entry combines the low and high resolution results instead of treating them as separate learning objectives.
The separate low/high resolution reconstructions obtain the best performance and robustness to changes in scale.

\section{Discussion}
\label{sec:discussion}

In this section, we share observations about \ours, sketch our vision for future work, and discuss high-level questions about \ours.

\paragraph{Computational complexity.} \ours requires a much smaller decoder than vanilla MAE---instead of a decoder depth of eight, \ours works well with a depth of three. In fact, with 322.9M vs 329.5M parameters using ViT-Large, \ours is smaller than vanilla MAE. However, GPU memory usage for equal batch sizes are higher for \ours since we reconstruct a higher resolution image in the \ours Decoder.

\paragraph{Multi-spectrality and modality.} Electro-optical (EO) satellites, such as the ones comprising the datasets mentioned in this work, capture light at different wavelengths. Each wavelength has a different sensor, and each sensor can have a different resolution. \ours requires input tensors to be stacked to pass through the model. This means that we are unable to use \ours when the input image's bands are all of different GSDs. Additionally, synthetic aperture radar (SAR) imagery is another form of remote sensing where resolution varies across a single band. Extending \ours to work with different resolution bands and modalities is reserved for future work.


\paragraph{Can the \ours methodology be applied to other backbones?} Methods such as ConvNeXt~\cite{liuConvNet2020s2022} provide competitive performance compared to Transformers. The core components of our work can be integrated, with additional work, into different architectures. The Laplacian Decoder in \ours can be engineered to ingest convolutional feature maps. Existing work on scale-aware CNNs can be extended to work with the Laplacian Decoder.

\begin{table}[t]
\centering
\begin{tabular}{c*{4}{ccc}}
 \textbf{Decoding Layers} & \textbf{KNN 50\%} & \textbf{KNN 100\%}\\\toprule
1& 76.0 & 78.4 \\\hline
2& 77.9  & 80.4 \\\hline
3& \textbf{78.1} & \textbf{80.7} \\\hline
4& 77.5  & 80.0 \\\hline
8& 77.7  & 78.9 \\\hline
\end{tabular}
\vspace{0.3cm}
\caption{Ablation results indicating that fewer transformer layers in the decoding stage tend to work better for \ours as determined by a KNN classification on RESISC-45 at a relative GSD of 50\% and 100\% of its native GSD.}
\label{tab:numlay}
\end{table}

\paragraph{Evaluating on more remote sensing datasets.} The field of remote sensing has had a renaissance in the last five years with the amount of available datasets. These can be generic, like Functional Map of the World, to highly specific, such as identifying illegal airstrips in Brazil~\cite{brazil_illegal_airstrips, chenFastAutomaticAirport2018} or identifying illegal fishing vessels~\cite{paoloXView3SARDetectingDark2022}. In fact, there are so many small, specific remote sensing datasets that entire review papers are written to enumerate them~\cite{xiongEarthNetsEmpoweringAI2022}. We chose to focus datasets with \textit{properties} of remote sensing that are relevant to multiscale representation learning.

\begin{table}
\centering
\resizebox{\columnwidth}{!}{%
\begin{tabular}{{ccccc}}
\textbf{Low Res} & \textbf{High Res} &  \textbf{Combined} & \textbf{KNN 50\%} & \textbf{KNN 100\%}\\\toprule
 & &\Checkmark & 77.6 & 80.2 \\\hline
 \Checkmark& &  &72.9  & 74.3 \\\hline
 &\Checkmark  & & 77.2 & 80.3 \\\hline
 \Checkmark&\Checkmark & & \textbf{78.1} & \textbf{80.7} \\\bottomrule
\end{tabular}
}
\vspace{0.3cm}
\caption{These ablation results indicate that reconstructing both the low resolution and high resolution components lead to robust performance. Note: when the high resolution component is reconstructed, the low-resolution residual is not used---the high resolution result is directly reconstructed. The ``Combined'' entry merges the low and high resolution results instead of treating them as separate losses. The evaluations are a kNN classification ($k$=20) on RESISC-45 at relative GSDs 50\% and 100\% of its native GSD.}
\label{tab:resolution}
\end{table}

\section{Conclusion}
Remote sensing imagery has accelerated the rate of scientific discovery in a broad set of disciplines. With increasingly precise methods to extract environmental indicators using computer vision methods, automated understanding of remotely sensed sources has become a mainstay in scientific literature. Remote sensing payloads are diverse and capture data at a wide range of resolutions, a feature heavily utilized by scientists. Current computer vision methods for remote sensing necessitate the training of a new model per input resolution. Not only is the training process expensive, but the overhead of curating a dataset at multiples scales makes this a daunting task.

We introduce \ours, a pretraining framework which introduces scale invariance into encoders that are used for a diverse set of downstream tasks. Our insights into scale-inclusive positional encodings and progressive multi-frequency feature extraction result in models that perform significantly better than state-of-the-art pretraining methods across (1) multiple scales and (2) many benchmarks.

Our goal is to take the extremely diverse and rich source of information present in remote sensing imagery and make it simple to use with minimal training iterations required. With the introduction of \ours, we hope to further accelerate the rate at which scientific disciplines create impact.

\section*{Acknowledgements}
We deeply thank Kyle Michel from Meta for providing us with his help during our time of need.
Satellite imagery and derived images used in this paper in are from datasets which redistribute imagery from Google Earth, DigitalGlobe, and Copernicus Sentinel 2022 data. 
Trevor Darrell's group was supported in part by funding from the Department of Defense as well as BAIR’s industrial alliance programs.
Ritwik Gupta is supported by the National Science Foundation under Grant No. DGE-2125913.






{\small
\bibliographystyle{ieee_fullname}
\bibliography{multiscale-mae,kitware}
}

\clearpage
\onecolumn
\appendix
\section{Datasets}
\label{sec:supp-datasets}

In our experiments, we used a total of ten datasets (\Cref{tab:supp-dataset-full}) for the tasks of land-use/land-cover classification and semantic segmentation. There are a large amount of remote sensing datasets in existence. Many remote sensing datasets fundamentally capture the same data with minor changes in location or distribution. We selected datasets with key, representative \textit{properties}. These properties include (1) a diversity in the amount of kinds of classes/objects represented, (2) a large spectrum of ground sample distances from (ideally) known sensor configurations, and (3) pansharpened, othrorectified, and quality controlled imagery and labels. We capture these properties in \Cref{tab:supp-dataset-full}. 

\subsection{Diversity in classes}
For both pretraining and downstream evaluations, it is a desirable property to include as much geographic and class diversity as possible. In order to capture a wide amount of classes in remote sensing, it is necessary to include multiple localities and environments. This property serves as a proxy for the amount of unique ``features" available in the dataset.

\begin{table*}[htbp]
  \centering
    \begin{tabular}{cccccc}
    \textbf{Dataset} & \textbf{Resolution (px)} & \textbf{GSD (m)} & \textbf{Number of Images} & \textbf{Number of Classes} & \textbf{Task Type} \\\toprule
    AiRound \cite{machadoAiRoundCVBrCTNovel2021} & 500   & 0.3 - 4800 & 11,753 & 11    & C \\\midrule
    CV-BrCT \cite{machadoAiRoundCVBrCTNovel2021} & 500   & 0.3 - 4800 & 24,000 & 9     & C \\\midrule
    EuroSAT \cite{helberIntroducingEurosatNovel2018} & 64    & 10    & 27,000 & 10    & C \\\midrule
    MLRSNet \cite{qiMLRSNetMultilabelHigh2020} & 256   & 0.1 - 10 & 109,161 & 46    & C \\\midrule
    Optimal-31 \cite{wangSceneClassificationRecurrent2019} & 256   & 0.5 - 8 & 1,860 & 31    & C \\\midrule
    RESISC-45 \cite{chengRemoteSensingImage2017} & 256   & 0.2 - 30 & 31,500 & 45    & C \\\midrule
    UC Merced \cite{yangBagofvisualwordsSpatialExtensions2010} & 256   & 0.3   & 2,100 & 21    & C \\\midrule
    WHU-RS19 \cite{daiSatelliteImageClassification2011} & 256   & 0.5   & 1050   & 19    & C \\\midrule
    fMoW \cite{christieFunctionalMapWorld2018} & Various & 0.3   & 1,047,691 & 62    & C \\\midrule
    SpaceNet v1 \cite{vanettenSpaceNetRemoteSensing2019} & Various & 0.5   & 6,940 & 2     & SS \\\bottomrule
    \end{tabular}%
    \vspace{0.3cm}
    \caption{Statistics of all datasets used in our experiments. Task types are classification (C) 
    and semantic segmentation (SS).}
    \label{tab:supp-dataset-full}%
\end{table*}%

\subsection{Spectrum of GSDs}
\ours is built to be invariant to the input absolute scale of the dataset. Many datasets are collected from a single sensor and processed in a uniform fashion. To validate that our method works with many resolutions, we included datasets which are collected from a variety of sensors but then processed in a uniform fashion. This excludes differences in processing as a factor affecting our experiments and narrowly targets resolution instead.

\begin{table*}[htbp]
\centering

\begin{tabular}{c|r||ccc||ccc||ccc||}
\multicolumn{1}{r}{}                 &                                    & \multicolumn{3}{c||}{$k=20$}                                                                & \multicolumn{3}{c||}{$k=100$}                                                       & \multicolumn{3}{c||}{$k=5$}                                                                  \\ 
\cmidrule{3-11}
\multicolumn{1}{c}{\textbf{Dataset}} & \multicolumn{1}{c||}{\textbf{Res}} & \textbf{Scale.} & \textbf{Sat.}  & \textbf{Conv.}                                           & \textbf{Scale.} & \textbf{Sat.}   & \textbf{Conv.}       & \textbf{Scale.} & \textbf{Sat.}  & \textbf{Conv.} \\ 
\toprule     
\multirow{6}{*}{AiRound}             & 16                                 & 0.401           & 0.375          & \textbf{0.423}                                           & 0.396           & 0.367           & \textbf{0.401 }      & 0.370           & 0.355          & \textbf{0.403} \\
                                     & 32                                 & \textbf{0.561}  & 0.510          & 0.539                                                    & \textbf{0.536}  & 0.491           & 0.517                & \textbf{0.541}  & 0.492          & 0.539          \\
                                     & 64                                 & \textbf{0.689}  & 0.607          & 0.658                                                    & \textbf{0.643}  & 0.579           & 0.621                & \textbf{0.692}  & 0.604          & 0.666          \\
                                     & 128                                & \textbf{0.743}  & 0.650          & 0.681                                                    & \textbf{0.690}  & 0.600           & 0.622                & \textbf{0.749}  & 0.660          & 0.690          \\
                                     & 256                                & \textbf{0.729}  & 0.662          & 0.658                                                    & \textbf{0.678}  & 0.621           & 0.602                & \textbf{0.731}  & 0.663          & 0.676          \\
                                     & 496                                & \textbf{0.670}  & 0.664          & 0.620                                                    & 0.609  & \textbf{0.613}           & 0.566                & \textbf{0.685}  & 0.669          & 0.632          \\ 
\midrule     
\multirow{6}{*}{CV-BrCT}             & 16                                 & 0.522           & 0.478          & \textbf{0.567}                                           & 0.485           & 0.443           & \textbf{0.513}       & 0.524           & 0.475          & \textbf{0.585} \\
                                     & 32                                 & 0.653  & 0.615          & \textbf{0.656}                                                    & 0.588           & 0.560           & \textbf{0.592}       & 0.695           & 0.644          & \textbf{0.699} \\
                                     & 64                                 & \textbf{0.744}  & 0.701          & 0.711                                                    & \textbf{0.674}  & 0.635           & 0.644                & \textbf{0.780}  & 0.727          & 0.754          \\
                                     & 128                                & \textbf{0.763}  & 0.725          & 0.732                                                    & \textbf{0.710}  & 0.662           & 0.667                & \textbf{0.805}  & 0.758          & 0.782          \\
                                     & 256                                & \textbf{0.761}  & 0.725          & 0.727                                                    & \textbf{0.694}  & 0.666           & 0.664                & \textbf{0.802}  & 0.770          & 0.771          \\
                                     & 496                                & \textbf{0.737}  & 0.727          & 0.709                                                    & 0.656  & \textbf{0.657}           & 0.631                & \textbf{0.792}  & 0.771          & 0.765          \\ 
\midrule     
\multirow{3}{*}{EuroSAT}             & 16                                 & 0.744           & 0.727          & \textbf{0.826}                                           & 0.699           & 0.695           & \textbf{0.788}       & 0.751           & 0.729          & \textbf{0.835} \\
                                     & 32                                 & \textbf{0.901}  & 0.876          & 0.898                                                    & \textbf{0.869}  & 0.854           & 0.863                & \textbf{0.912}  & 0.871          & 0.909          \\
                                     & 64                                 & \textbf{0.956}  & 0.931          & 0.940                                                    & \textbf{0.935}  & 0.913           & 0.914                & \textbf{0.960}  & 0.934          & 0.947          \\ 
\midrule     
\multirow{5}{*}{MLRSNet}             & 16                                 & 0.563           & 0.491          & \textbf{0.607}                                           & 0.535           & 0.461           & \textbf{0.549}       & 0.551           & 0.479          & \textbf{0.617} \\
                                     & 32                                 & \textbf{0.772}  & 0.677          & 0.744                                                    & \textbf{0.726}  & 0.625           & 0.688                & \textbf{0.772}  & 0.684          & 0.762          \\
                                     & 64                                 & \textbf{0.893}  & 0.815          & 0.851                                                    & \textbf{0.849}  & 0.754           & 0.792                & \textbf{0.911}  & 0.839          & 0.876          \\
                                     & 128                                & \textbf{0.936}  & 0.875          & 0.894                                                    & \textbf{0.892}  & 0.814           & 0.834                & \textbf{0.950}  & 0.899          & 0.918          \\
                                     & 256                                & \textbf{0.918}  & 0.892          & 0.882                                                    & \textbf{0.862}  & 0.840           & 0.817                & \textbf{0.940}  & 0.913          & 0.910          \\ 
\midrule     
\multirow{5}{*}{OPTIMAL-31}          & 16                                 & 0.354           & 0.322          & \textbf{0.439}                                           & 0.312           & 0.298           & \textbf{0.370}       & 0.317           & 0.319          & \textbf{0.418} \\
                                     & 32                                 & 0.574  & 0.500          & \textbf{0.587}                                                    & \textbf{0.567}  & 0.508           & 0.545                & \textbf{0.565}  & 0.519          & 0.561          \\
                                     & 64                                 & \textbf{0.793}  & 0.609          & 0.698                                                    & \textbf{0.742}  & 0.561           & 0.598                & \textbf{0.782}  & 0.646          & 0.688          \\
                                     & 128                                & \textbf{0.816}  & 0.670          & 0.714                                                    & \textbf{0.731}  & 0.646           & 0.595                & \textbf{0.809}  & 0.694          & 0.725          \\
                                     & 256                                & \textbf{0.739}  & 0.681          & 0.646                                                    & \textbf{0.653}  & 0.638           & 0.550                & \textbf{0.761}  & 0.731 & 0.693                   \\ 
\midrule     
\multirow{5}{*}{RESISC}              & 16                                 & 0.382           & 0.347          & \textbf{0.458}                                           & 0.370           & 0.327           & \textbf{0.428}       & 0.353           & 0.323          & \textbf{0.435} \\
                                     & 32                                 & \textbf{0.628}  & 0.527          & 0.601                                                    & \textbf{0.597}  & 0.505           & 0.568                & \textbf{0.609}  & 0.508          & 0.592          \\
                                     & 64                                 & \textbf{0.798}  & 0.667          & 0.731                                                    & \textbf{0.754}  & 0.631           & 0.677                & \textbf{0.803}  & 0.667          & 0.734          \\
                                     & 128                                & \textbf{0.864}  & 0.748          & 0.798                                                    & \textbf{0.819}  & 0.699           & 0.743                & \textbf{0.882}  & 0.762          & 0.817          \\
                                     & 256                                & \textbf{0.826}  & 0.758          & 0.762                                                    & \textbf{0.761}  & 0.708           & 0.690                & \textbf{0.850}  & 0.771          & 0.788          \\ 
\midrule     
\multirow{5}{*}{UC Merced}           & 16                                 & 0.524           & 0.472          & \textbf{0.598}                                           & 0.400           & 0.370           & \textbf{0.462}       & 0.512           & 0.488          & \textbf{0.617} \\
                                     & 32                                 & \textbf{0.767}  & 0.670          & 0.683                                                    & \textbf{0.605}  & 0.535           & 0.593                & \textbf{0.828}  & 0.682          & 0.726          \\
                                     & 64                                 & \textbf{0.842}  & 0.795          & 0.771                                                    & 0.719  & \textbf{0.729}           & 0.652                & \textbf{0.884}  & 0.842          & 0.845          \\
                                     & 128                                & \textbf{0.858}  & 0.788          & 0.750                                                    & 0.662           & \textbf{0.738}  & 0.655                & \textbf{0.884}  & 0.847          & 0.838          \\
                                     & 256                                & 0.762           & \textbf{0.802} & 0.700                                                    & 0.595           & \textbf{0.757}  & 0.590                & \textbf{0.851}  & 0.842          & 0.817          \\ 
\midrule     
\multirow{5}{*}{WHU-RS19}            & 16                                 & 0.545           & 0.445          & \textbf{0.576}                                           & 0.400           & 0.380           & \textbf{0.562}       & 0.525           & 0.490          & \textbf{0.631} \\
                                     & 32                                 & 0.650           & \textbf{0.729} & 0.670                                                    & 0.610           & \textbf{0.675}  & 0.576                & \textbf{0.760}  & 0.690          & 0.754          \\
                                     & 64                                 & \textbf{0.850}  & 0.805          & 0.833                                                    & \textbf{0.770}  & 0.730           & 0.680                & \textbf{0.920} & 0.840           & 0.837          \\
                                     & 128                                & \textbf{0.970}  & 0.910          & 0.882                                                    & \textbf{0.890}  & \textbf{0.890}  & 0.685                & \textbf{0.985} & 0.895           & 0.941          \\
                                     & 256                                & \textbf{0.960}  & 0.940          & 0.892                                                    & 0.880           & \textbf{0.925}  & 0.709                & \textbf{0.975} & 0.945           & 0.931          \\
\bottomrule
\end{tabular}%
\\ 
    \vspace{0.3cm}
    \caption{\textbf{\ours outperforms SatMAE and ConvMAE on kNN classification across a variety of $k$, across a variety of resolutions.} kNN Classification results for \ours, SatMAE and ConvMAE across a variety of $k$. Resolution is reported in pixels.}
    \label{tab:supp-knn-full}%
\end{table*}

\subsection{Quality control}
It is hard to assess the quality of remote sensing datasets without manually verifying a majority of instances of the data. We mandated that images used are pansharpened (and therefore the highest resolution possible to extract from the sensor), orthorectified (and therefore well-aligned with the geodetic ellispoid), and projected to the same coordinate reference system. This eliminates large differences in sensor-to-image processing.

\section{Laplacian and Upsampling Block Architectures}
\begin{figure}
    \centering
    \includegraphics[width=0.7\textwidth]{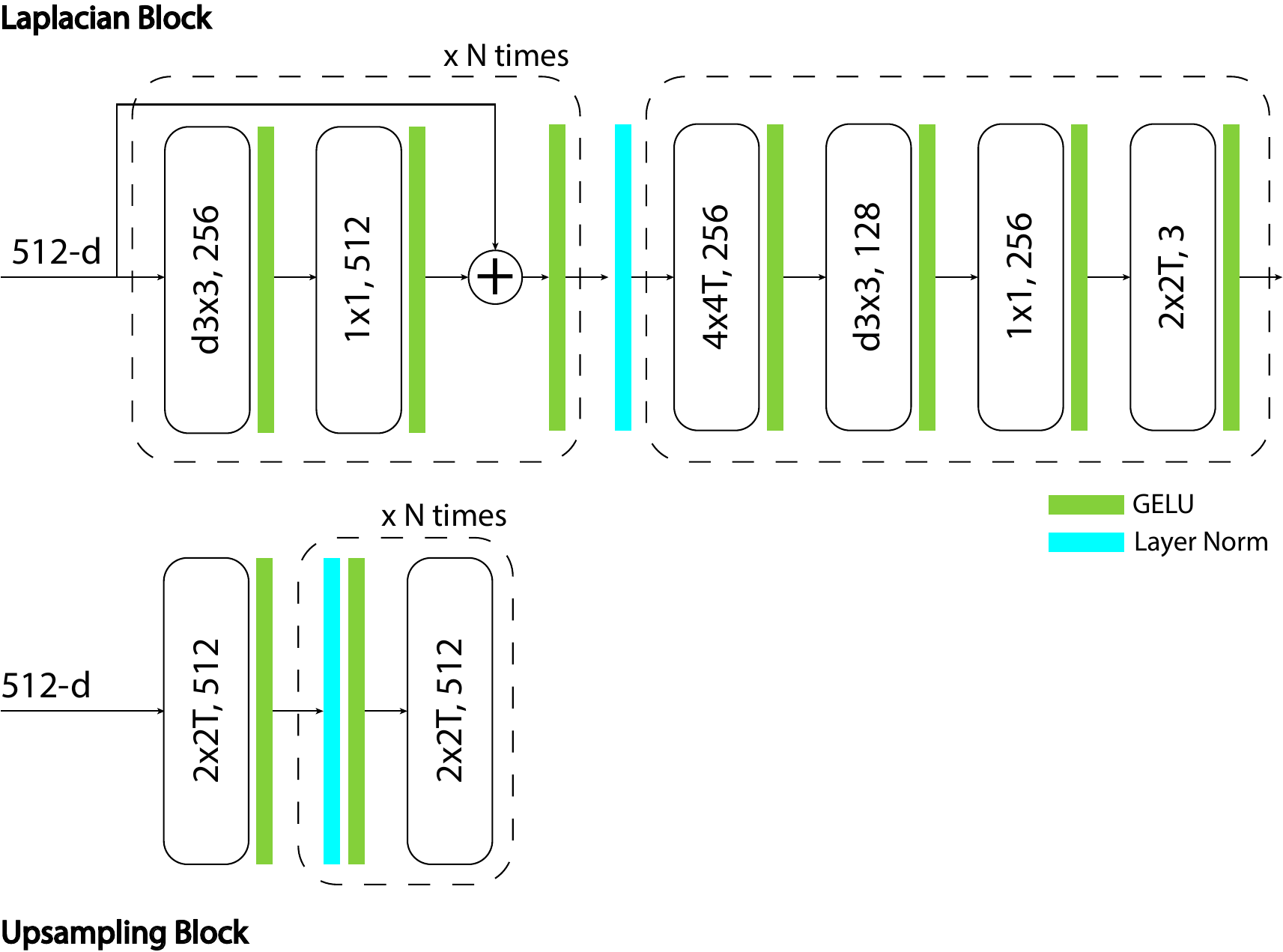}
    \caption{(top) The Laplacian Block (LB) is a fully convolutional architecture consists of a chain of Feature Mapping Block followed by one final Reconstruction Block. (bottom) The UpSampling Block (UB) consists of a series of transpose convolution layers separated by LayerNorm and GELU activation. }
    \label{fig:supp-lb-upsample}
\end{figure}

\Cref{fig:supp-lb-upsample} illustrates the architecture of Laplacian and Upsampling block architectures described below.
\subsection{Laplacian Block}
Laplacian Blocks are used to reconstruct the target at a specific resolution and frequency. A Laplacian Block consists of a chain of Feature Mapping Block, which distills information at a specific frequency, followed by one final Reconstruction Block, which generates the final output. A Feature Mapping Block consists of a 3x3 depth-wise convolution layer with GELU activation, followed by 1x1 convolution.  A Reconstruction Block consists of a 4x4 transpose convolution layer followed by a 3x3 depth-wise convolution layer, a 1x1 convolution layer, and a 2x2 transpose convolution layer. In our experiments, we have two Feature Mapping Blocks per Laplacian Block.

\subsection{Upsampling Block}
Upsampling Blocks are used to upsample the feature map to a higher resolution. It consists of a series of 2x2 transpose convolution layers with LayerNorm and GELU activation between them. The number of such transposed convolution layers are a function of the output and input resolution. This is a progressive process in which we repetitively upsample the feature map by a factor of 2 until we reach the desired target resolution. \Cref{fig:supp-lb-upsample} illustrates the architecture of these two blocks. 

\section{Evaluation Details}
As discussed in the main experimental section, we investigated the quality of representations learned from \ours pretraining through a set of experiments that explore their robustness to scale as well as their transfer performance to additional tasks. We provide more information and details on these evaluations here.
In order to compare with SatMAE~\cite{congSatMAEPretrainingTransformers2022} and ConvMAE~\cite{gaoConvMAEMaskedConvolution2022}, for our main experiments, we pretrained \ours with a ViT-Large model using the Functional Map of the World (FMoW) RGB training set, which consists of $363.6$k images of varying image resolution and GSD. 
The initial higher resolution image $I_{\text{hr}}$ is taken as a random 448px$^2$ crop of the input image, and the input image $I$ is then a downsampled 224px$^2$ from $I_{\text{hr}}$.
The low frequency groundtruth is obtained by downscaling $I_{\text{hr}}$ to 14px$^2$ and then upscaling to 224px$^2$, while the high frequency groundtruth is obtained by downscaling $I_{\text{hr}}$ to 56px$^2$ and then upscaling to 448px$^2$ and subtracting this image from $I_{\text{hr}}$.
This is a common method for band pass filtering used in several super resolution works, where a high to low to high resolution interpolation is used to obtain only low frequency results, and then high frequency results are obtained by subtracting the low frequency image.

As further discussed in the main experimental section, we evaluate the quality of representations from \ours by freezing the encoder and performing a nonparametric k-nearest-neighbor (kNN) classification with eight different remote sensing imagery classification datasets with different GSDs, none of which were encountered during pretraining. All kNN evaluations were conducted on 4 GPUs. Results are in \Cref{tab:supp-knn-full}.
The kNN classifier operates by encoding all train and validation instances, where each embedded instance in the validation set computes the cosine distance with each embedded instance in the training set, where the instance is classified correctly if the majority of its k-nearest-neighbors are in the same class as the validation instance. The justification for a kNN classifier evaluation is that a strong pretrained network will output semantically grouped representation for unseen data of the same class. This evaluation for the quality of representations occurs in other notable works~\cite{caronEmergingPropertiesSelfSupervised2021, chenExploringSimpleSiamese2021, wuUnsupervisedFeatureLearning2018}.

\section{Visualization of SpaceNet Segmentation}
\begin{figure}
    \includegraphics[width=\textwidth]{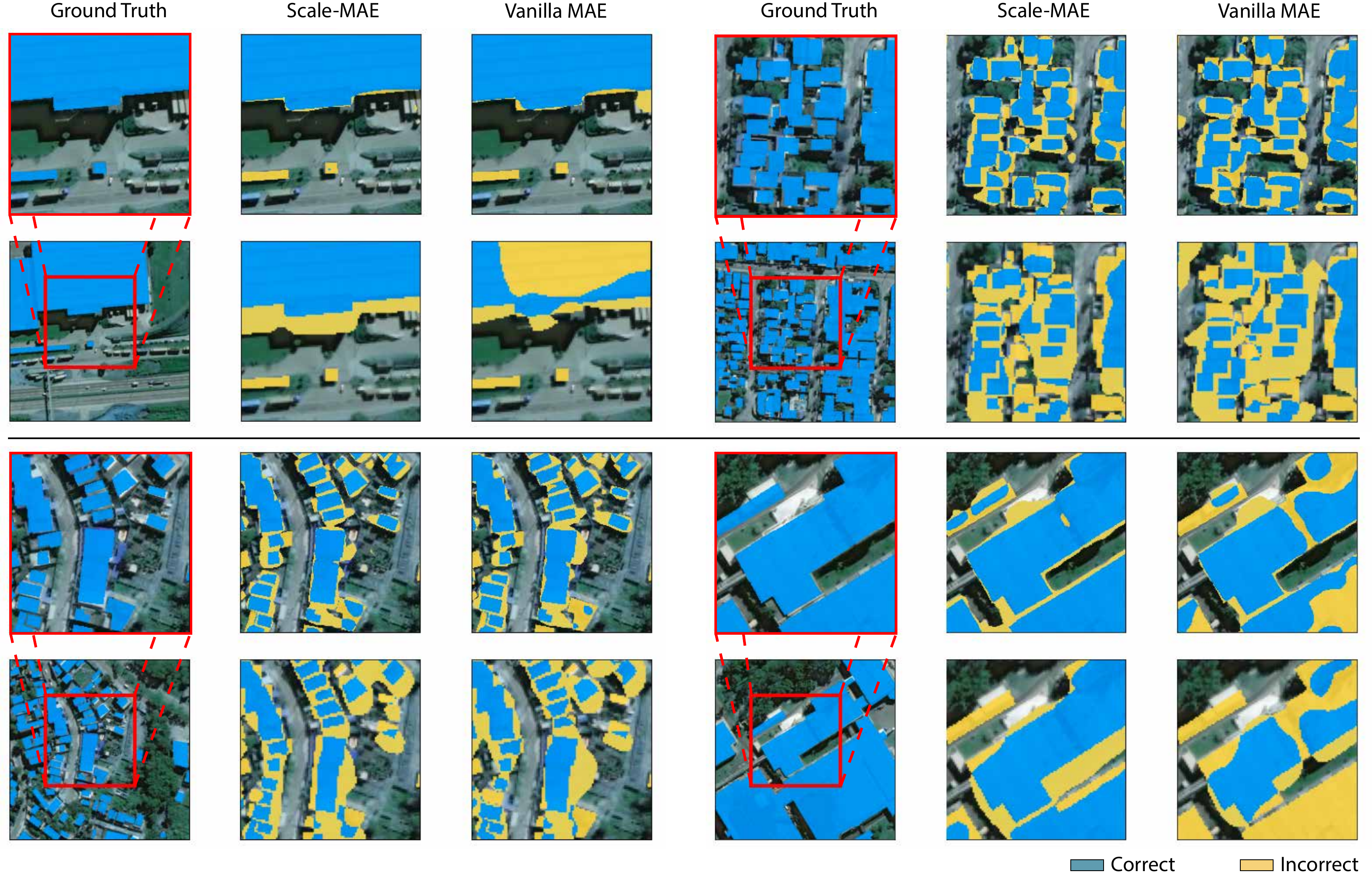}
    \caption{Visualization of Segmentation Results on SpaceNet. The left, center, right columns are ground truth labels, \ours and vanilla MAE, respectively. The top row shows a 0.3m GSD image and the bottom row shows a 3.0m GSD image. As shown in the figure, \ours performs better at both higher and lower GSDs.}
    \label{fig:supp-lb-upsample2}
\end{figure}

\Cref{fig:supp-lb-upsample2} shows an additional set of segmentation examples comparing \ours and vanilla MAE pre-trained on FMoW and finetuned on SpaceNet v1. The left, center, right columns are ground truth labels, \ours and vanilla MAE respectively. The top row shows a 0.3m GSD image and the bottom row shows a 3.0m GSD image. As shown in the figure, \ours performs better at both higher and lower GSDs.

 \section{Glossary}
 \subsection{Ground sample distance}
Ground sample distance (GSD) is the distance between the center of one pixel to the center of an adjacent pixel in a remote sensing image. GSD is a function of sensor parameters (such as its dimensions and focal length), image parameters (the target dimensions of the formed image), and the geometry of the sensor with respect to the object being imaged on the Earth. Remote sensing platforms frequently have multiple sensors to capture different wavelengths of light. Each of these sensors have varying parameters, resulting in different GSDs for an image of the same area. Additionally, the ground is not a uniform surface with changes in elevation common across the swath of the sensor. In total, a remote sensing platform has a sense of absolute scale that varies along two dimensions: (1) spectrally depending on the sensor used to capture light, and (2) spatially depending on surface elevation.

\end{document}